\documentclass[multiauthors,onecolumn,cfinal]{LaTeXclass/cohere}

\usepackage{lipsum}
\usepackage{pdfpages}
\usepackage{colortbl}
\usepackage{tabulary}
\usepackage{longtable}
\usepackage{array}
\usepackage{placeins}
\usepackage{tablefootnote}
\usepackage{tcolorbox}
\usepackage{wrapfig}
\usepackage{breqn} 

\usepackage{pifont}
\definecolor{deeppurple}{HTML}{9e02f7}
\definecolor{forestgreen}{HTML}{2e7d43}
\usepackage{multirow}
\usepackage{tabularx}

\usepackage[utf8]{inputenc}

\usepackage{arydshln}

\usepackage{xspace} 
\usepackage{makecell} 
\usepackage{multirow}

\usepackage[framemethod=TikZ]{mdframed}
\usepackage{multicol}
\usepackage{ragged2e}
\usepackage{nicematrix}
\usepackage{comment}
\usepackage{amssymb}
\newcommand{\cmark}{\ding{51}} 
\newcommand{\xmark}{\ding{55}} 

\setcitestyle{number,square}

\title{One Tokenizer To Rule Them All: \\Emergent Language Plasticity via Multilingual Tokenizers}

\author{name={Diana Abagyan\fa},affiliation={1}}
\author{name={Alejandro R. Salamanca},affiliation={1}}
\author{name={Andres Felipe Cruz-Salinas},affiliation={2}}
\author{name={Kris Cao},affiliation={2}}
\author{name={Hangyu Lin},affiliation={2}}
\author{name={Acyr Locatelli},affiliation={2}}
\author{name={Marzieh Fadaee},affiliation={1}}
\author{name={Ahmet Üstün\psa},affiliation={1}}
\author{name={Sara Hooker\psa},affiliation={1}}
\affiliations{
    \item[1] Cohere Labs
    \item[2] Cohere
}

\corresponding[*]{\{\url{dianaabagyan}, \url{ahmet}, \url{sarahooker}\}\url{{@cohere.com}}}

\abstract{
\justifying
Pretraining massively multilingual Large Language Models (LLMs) for many languages at once is challenging due to limited model capacity, scarce high-quality data, and compute constraints. Moreover, the lack of language coverage of the tokenizer makes it harder to address the gap for new languages purely at the post-training stage. In this work, we study what relatively cheap interventions early on in training improve ``language plasticity'', or adaptation capabilities of the model post-training to new languages. We focus on tokenizer design and propose using a \textit{universal} tokenizer that is trained for more languages than the primary pretraining languages to enable efficient adaptation in expanding language coverage after pretraining. Our systematic experiments across diverse groups of languages and different training strategies show that a universal tokenizer enables significantly higher language adaptation, with up to 20.2\% increase in win rates compared to tokenizers specific to pretraining languages. Furthermore, a universal tokenizer also leads to better plasticity towards languages that are completely unseen in the tokenizer and pretraining, by up to 5\% win rate gain. We achieve this adaptation to an expanded set of languages with minimal compromise in performance on the majority of languages included in pre-training. 
}

\renewcommand{\FONTauthors}{\bfseries\large}

\begin{document}

\begin{figure}[!t]
    \centering
    \includegraphics[width=.5\linewidth]{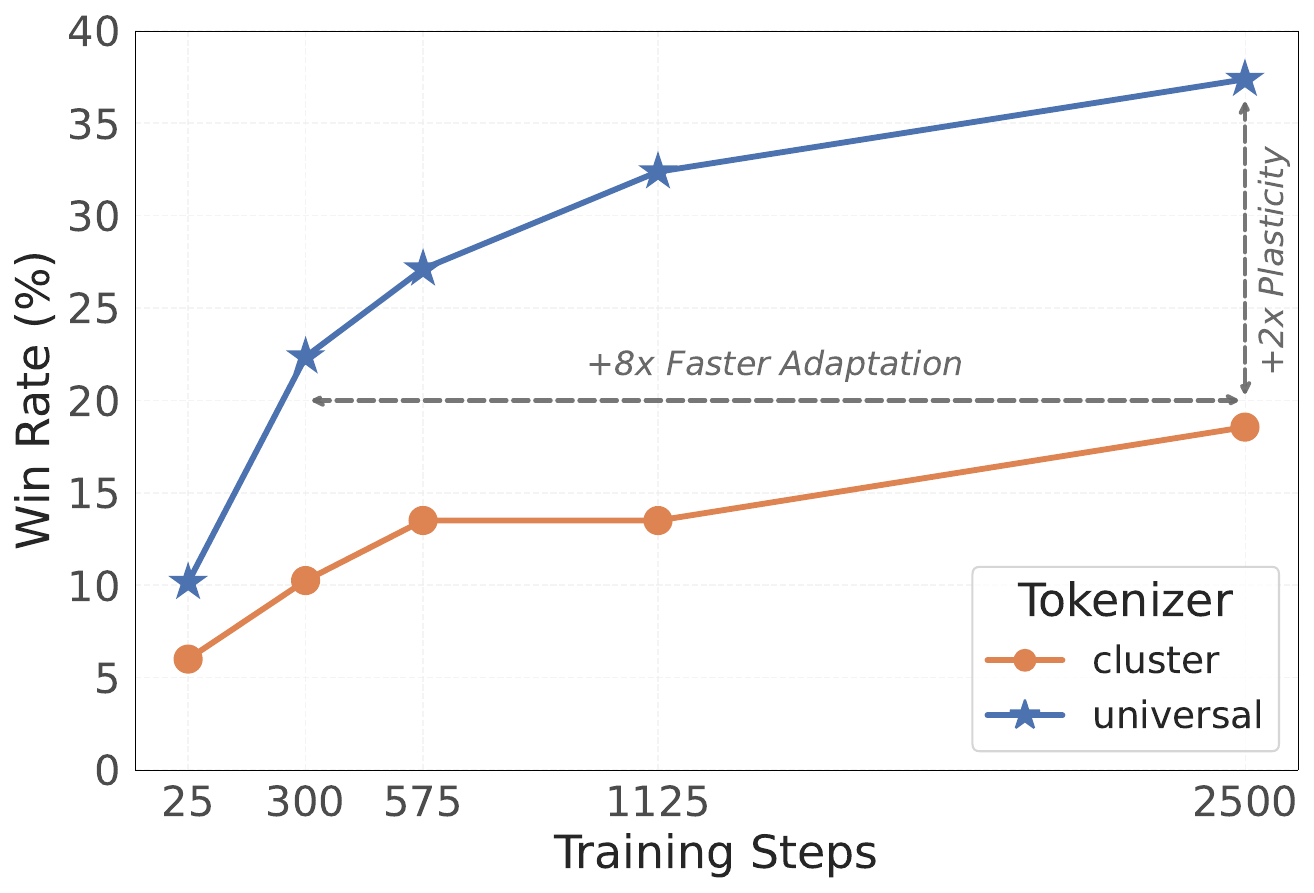}
    \caption{\textsc{Universal} tokenizer exhibits \textbf{+2x higher plasticity} with \textbf{+8x faster adaptation} compared to the cluster-specific baseline tokenizer. Average win rates on the expanded (new) language subset during continued pretraining that involves both primary and expanded language subsets (\S~\ref{sec:adaptation-strategies}).}
    \label{fig:adaptation-speed} 
\end{figure}

\section{Introduction}

There are only a handful of research labs with enough compute resources and expertise to train large AI systems at scale \citep{stanford_ai_index_2025,Hooker2024OnTL}.
Most researchers and practitioners are forced to choose among available pretrained models for downstream tasks, even if they are not tailored to their use cases. Nowhere is this tension more evident than in the multilingual setting \citep{joshi-etal-2020-state,singh2024aya,ustun-etal-2024-aya}, where limited investment in multilingual support in pretraining often results in significant gaps in language coverage in state-of-the-art LLMs \citep{holtermann-etal-2024-evaluating}. 

This imbalance in language coverage has created a growing divide in the cost of use for particular language users as marginalized languages require more tokens and incur higher latency for generations \citep{ji2023betterinstructionfollowinglanguage,cui2023efficient, ahia-etal-2023-languages}, restricting speakers of low-performing languages to lower quality technology \citep{held2023materiallenscolonialitynlp, durmus2024measuringrepresentationsubjectiveglobal,nicholas2023lost,ojo2025afrobenchgoodlargelanguage}. Further compounding these issues, once a model is pretrained, it is hard to steer towards new behavior using post-training alone \citep{wang-etal-2025-language}. Unless the tokenizer has been calibrated to a new language during training, it often requires far more significant amount of data and intricate optimization steps \citep{muller-etal-2021-unseen}. 

\textbf{Multilingual plasticity} represents the capability of the language model to quickly adapt to lingual distribution shifts to the downstream target, which in our case, involves a new set of focus languages \citep{chen2023}. Given that pretraining requires the bulk of compute and cost resources, any intervention made at this stage that improves the \textit{plasticity} for downstream developers and researchers is beneficial. 

In this work, we investigate minimal and efficient pretraining interventions to reduce later adaptation costs.
In particular, we identify tokenization as an area with relatively low cost of intervention, but potential for large downstream gains. We ask: \textit{Can we leverage tokenizers with broad language coverage to improve the plasticity of LLMs without hurting pretraining performance?}

We hypothesize that a universal tokenizer that is trained on more languages than the primary pretraining languages, introduced from the start of pretraining, enables quick and effective interventions for adapting a model to new languages. This significantly diverges from the previous work that focuses on techniques such as vocabulary extension \citep{wang-etal-2020-extending} or retraining the embedding layer \citep{artetxe-etal-2020-cross} after pretraining. These techniques are more costly, needing more resources like training budget, with varying degrees of success across languages \citep{limisiewicz-etal-2023-tokenization,sharthak2025achievingtokenizerflexibilitylanguage,nag-etal-2025-efficient}.

\begin{figure*}[t]
    \centering
    \includegraphics[width=\linewidth]{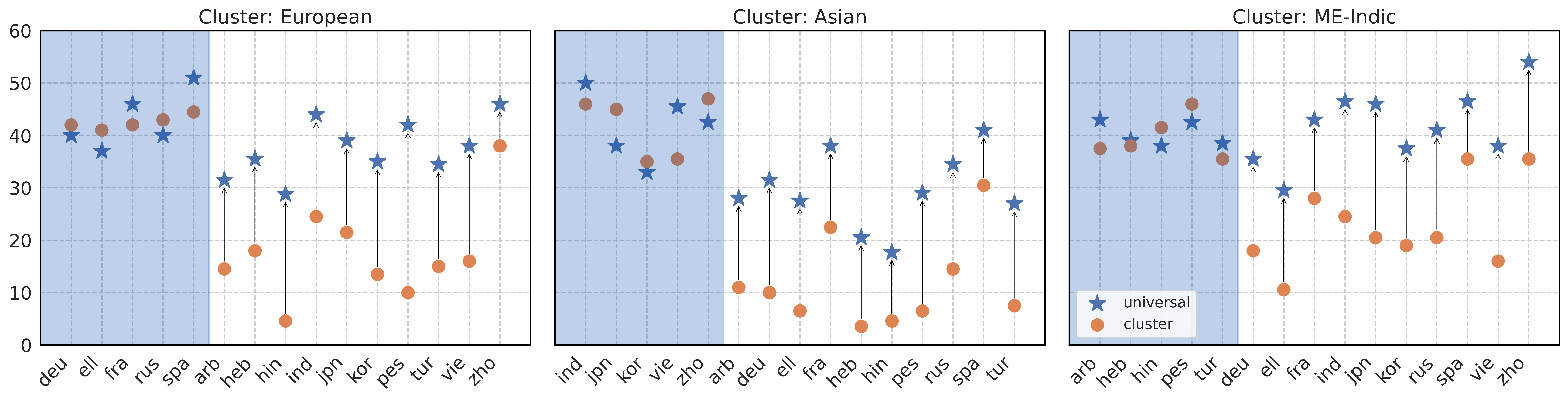}
    \caption{Win rates for models trained with the \textsc{Universal} and \textsc{Cluster} tokenizers against Dolly generations \citep{singh2024aya}. The shaded blue portion represents a subset of primary languages, and the white represents expanded languages. On average, \textsc{Universal} tokenizer achieves an increase of 18.9\% win rate on expanded subset languages, and +0.3\% on primary languages compared to the \textsc{Cluster} tokenizers across clusters.}
    \label{fig:all-langs-cooldown} 
\end{figure*}

We systematically investigate the impact of multilingual tokenizers through an exhaustive set of ablations at pretraining scale, which requires a significant investment of resources. We vary tokenizer, language subsets and adaptation strategy across 69 languages, and we find the following results:
\begin{enumerate}
    \item The \textbf{\textsc{Universal}} tokenizer significantly improves adaptation to new (expanded) languages, achieving an average of 19\% higher win rate in continued pretraining experiments, compared to the baseline tokenizers specialized to pretraining languages. In addition to achieving higher adaptation, the \textbf{\textsc{Universal}} tokenizer exhibits almost the same performance on primary languages with no more than 2\% difference in downstream evaluation against the baseline tokenizer.
    \item For targeted adaptation where the new languages are the only focus, the \textbf{\textsc{Universal}} tokenizer achieves an average of 14.6\% improvement over the baseline tokenizer for the expanded languages subset. Furthermore, for adapting to fully \textbf{unseen} languages, not included both in the tokenizer and pretraining (the most extreme case of adaptation), \textsc{Universal} tokenizer outperforms the baseline by up to 5\% gain in win rates across 7 heavily under-resourced languages.  
    \item We find that the \textsc{Universal} tokenizer enables more than 8x faster adaptation performance, requiring much less additional training, and therefore minimal costs. We believe that this dramatically benefits practitioners who want to extend the language coverage of a pretraining model with minimal intervention. 
\end{enumerate}

\section{Methodology and Experimental Setup}

\subsection{Methodology and Core Ablations}
\label{sec:strategies}

\textbf{Language Coverage and Model Variants.} Our experiments include 62 typologically and lexicographically diverse languages, broken up into three geographically motivated clusters: (1) European languages, (2) Asian languages, and (3) Middle-Eastern and Indic languages (referred to as ME-Indic throughout the paper). For each geo-cluster, we pretrain language models primarily on the languages within that cluster (referred to as \textbf{primary} subset) and we use the remaining languages (referred to as \textbf{expanded} subset) as reference points for plasticity adaptation experiments. For example, for the European cluster, the primary subset consists of languages such as Spanish, Russian, and Portuguese, and the expanded subset includes the 10 languages outside of the dominant training data for that cluster. In addition, we also consider 7 \textbf{fully unseen} languages, such as Sinhala and Kazakh, which were not present in the tokenizer or base model training data. The full language list with clusters is provided in Appendix \ref{app:language-list}. 

\noindent\textbf{Adaptation Strategies.}\label{sec:adaptation-strategies} One of our goals is to introduce highly plastic and adaptable model properties. A practitioner may choose to conduct language adaptation using a variety of different training strategies, influenced by what data they may have available. Ideally, the choices we make for the tokenizer allow for improved plasticity given any approach taken after pretraining. Hence, we evaluate our interventions under various different adaptation strategies, which include continued pretraining with data in both \textbf{primary and expanded} language subsets, targeted adaptation for \textbf{expanded} languages, and targeted adaptation for \textbf{fully unseen} languages. We briefly describe both these strategies and experimental details below:

\begin{itemize}
    \item \textbf{Continued pretraining with data from primary and expanded languages:} The objective for this strategy is to increase language coverage of the model, so that it supports both primary and expanded languages.
    Half of the training mix consists of an even distribution of all languages in the instruction finetuning data, and the other half is a standard cooldown mix with high-quality datasets (See \S~\ref{sec:cooldown-data}). This imparts instruction-following abilities to our base models, and also allows for evaluation on both the primary and expanded languages.
    \item \textbf{Targeted adaptation (expanded languages):} In these experiments, we explore a targeted language adaptation through supervised fine-tuning. The post-training data solely consists of instruction-style data in the expanded language subsets for each cluster model. This allows us to isolate the effect of introducing new languages that are not focused on during pretraining but represented in the tokenizer.
    \item \textbf{Targeted adaptation (fully unseen languages):} In the final set of experiments, we explore the most extreme setting of targeted adaptation to fully unseen languages that are not seen in the tokenizer or pretraining. In this setting, we consider the availability of the data in one language only for each experiment, thus, we fine-tune the base model on one language at a time. This ablation enables evaluating our approach for adaptation under a heavily under-resourced scenario. 
\end{itemize}

\noindent\textbf{Tokenizer Variants.} We train a massively multilingual tokenizer using data from all 62 languages as well as cluster-specific tokenizers that represent only the primary language subsets. Throughout the paper, we refer to these tokenizers as \textbf{\textsc{Universal}} and \textbf{\textsc{Cluster}} tokenizers, respectively. We include more details about the tokenizer training in Section \ref{sec:tokenizer-training}. 

\subsection{Experimental Set-up}

\noindent\textbf{Pretraining datasets.}
Models are pretrained with a mixture of English, code, and multilingual corpora, where data weights are distributed as 55\%, 15\%, and 30\%, respectively. Upweighting English in multilingual training is a common practice, due to higher task coverage and quality, which is crucial for cross-lingual transfer \citep{dash2025ayavisionadvancingfrontier,singh2024aya,ravisankar2025mapenglishrolecrosslingual}. We also include code data as it has become a standard part of the training recipe even for natural language models, and has been found to boost performance on other tasks when included in pre-training \citep{matraining,aryabumi2024code}. We use a large corpus of data from a variety of public and proprietary sources. For models we pretrain with the \textsc{Universal} tokenizer, we reallocate 5\% of the training mixture from English data and uniformly distribute it among all the expanded languages, to avoid undertrained tokens in the vocabulary \citep{land-bartolo-2024-fishing}. However, in Section \ref{sec:expanded-subset-percentage}, we ablate this percentage and show that even when no expanded language subset data is included in pretraining, the \textsc{Universal} tokenizer significantly improves multilingual plasticity. 

\noindent\textbf{Cooldown and instruct datasets.}\label{sec:cooldown-data} For continued pretraining, we use cooldown data that involves upweighting higher quality datasets, comprised of text, math, code, and instruct-style data \citep{aryabumi2024code}. It has been found by recent work to improve performance on downstream tasks, in particular by helping impart instruction-following capabilities \citep{parmar2024reusedontretrainrecipe, geminiteam2025}. We include a high-quality mix of proprietary and open data, much of which was created by following a multilingual data arbitrage strategy \citep{odumakinde2024multilingualarbitrageoptimizingdata}, covering 100,000 prompt-completion pairs in 23 languages. Finally, for experiments on fully unseen languages, we emulate a realistic data-constrained regime, and use only 14,800 instructions per language from the translated Dolly training set from Aya Collection \citep{singh2024aya}.

\noindent\textbf{Training details.} 
For our experiments, as standard for most LLMs, we use the Transformer-based decoder-only architecture \citep{vaswani-2017-transformers, Radford2018ImprovingLU}. Our architecture includes key optimizations such as Parallel Attention Blocks \citep{chowdhery2023palm}, Grouped Query Attention \citep{ainslie-etal-2023-gqa}, SwiGLU activation function \citep{shazeer2020gluvariants}, and Rotary Positional Embeddings \citep{10.1016/j.neucom.2023.127063}.

Our ablations are extensive and require a large amount of pretraining runs. Given the huge amount of compute required for pretraining, where training a 3.3B parameter model on 128 Nvidia H100 GPUs takes 11 hours, we focus on only 3.3 billion parameter language models for ablations. We train each base variant for 100 billion tokens using a total of 25,000 steps. 
Given the number of experiments we run, and the variety of factors we evaluate, this model size and amount of training steps is at the edge of what is computationally feasible at pretraining scale. Overall, the goal is not to emulate the settings of a full pretraining run, but to get sufficient signal about the relative merits of different approaches. In the continued pretraining strategy, we train for an additional 10.5B tokens and the targeted adaptation are done for 4 epochs over the respective datasets for each experiment. Additional training and infrastructure details are provided in Appendix \ref{app:training-details}.

\noindent\textbf{Infrastructure.} 
We use Nvidia H100 GPUs for training and evaluation. We use FAX \citep{yoo2022scalable}, which is a high-performance training framework built on JAX \citep{jax2018github}, enabling efficient tensor and model parallelism.

\subsection{Tokenizer Training} \label{sec:tokenizer-training}

All tokenizers are trained using the Byte Pair Encoding algorithm \citep{sennrich-etal-2016-neural}. Additional implementation details about the tokenizer training is given in Appendix \ref{app:tokenizer-details}.

\noindent\textbf{Language weighting.} In addition to varying the coverage of the tokenizer, with some being trained on \textsc{Universal} and \textsc{Cluster} language coverage, we also invest in a methodology that adjusts the weighting based upon availability of data. In contrast to traditional approaches which sample uniformly across all data and end up dominated by most frequent languages, we consider two factors: (1) natural distribution of the data available across languages, and (2) language buckets formed by languages that share the same family and script (which are more likely to share tokens). Within each language bucket,  we use uniform weighting across languages. Concretely, for a language $i$, where $w_i^d$ and $w_i^b$ denote weights for data distribution and language bucket, respectively, we compute the language weights in the tokenizer data mixture as follows: 

\begin{equation}
\label{eq:tokenizer}
    w_i = \frac{w^d_{i} . w^b_{i}}{\sum_{n} w^d_{n} . w^b_{n}}
\end{equation}

This way, we balance natural data distribution (skewed through the high-resource languages) with language bucketing in a principled manner, ensuring that there is equitable representation for diverse scripts and lower-resourced languages. Our pretraining experiments (Section \ref{sec:pretraining-results}, Appendix \ref{app:compression-ratio}) show that our specialized weighting combining language bucketing with size-proportional data distribution enables better compression ratios than uniform weighting and achieves better downstream performance. For the remainder of this work, we use specialized weighting throughout experiments, unless specified otherwise. 

\noindent\textbf{Vocabulary size.} We use a vocabulary size of 250k tokens in our main experiments. However, in \S~\ref{sec:vocab-size}, we explore how vocabulary size impacts performance for both \textsc{Universal} and \textsc{Cluster} tokenizers, and vary the vocabulary size between 100k, 175k, and 250k to understand the impact.

\subsection{Evaluation}
\noindent\textbf{Open-ended evaluation.}
\citet{goldman-etal-2024-unpacking} find that generative tasks are more informative than classification in evaluating tokenizers, likely due to the number of generation steps. Following \citet{ustun-etal-2024-aya}, the quality of generations is assessed using LLM-as-a-Judge win rates, where original generations are used as the reference answer. We use the \texttt{dolly\_human\_edited} and the \texttt{dolly\_machine\_translated} splits of the Aya Evaluation Dataset \citep{singh2024aya} as test data for this task, which are formed by translating 200 held-out examples from the Dolly-15k \citep{DatabricksBlog2023DollyV2}. We use 15 adaptation languages for open-ended evaluation, listed in Appendix \ref{app:language-list}.

Prior work has shown that LLMs as evaluators are reasonable proxies and aligned with human preferences also in multilingual settings \citep{ustun-etal-2024-aya, singh2025leaderboardillusion, dang2024ayaexpansecombiningresearch,kreutzer2025d}. We use Command-A \citep{cohere2025commandaenterprisereadylarge} as the judge model, given its reported strength as the best open-weights judge on multilingual setting, scoring closely to GPT4o ~\citep{gureja2024m,pombal2025m}. The full judge prompt is included in Appendix \ref{app:winrate-prompt}. 

\noindent\textbf{Task-specific performance.} 
We use two task specific evaluations for multilingual evaluations. Belebele \citep{bandarkar-etal-2024-belebele} is a multiple-choice question machine-reading comprehension (MRC) dataset representing 122 language variants. Multilingual MMLU (M-MMLU) \citep{dac2023okapi} is a machine-translated version of the original MMLU dataset \citep{hendrycks2021mmlu} that contains questions ranging in topic from STEM to humanities. 

\noindent\textbf{English-only evaluation.} \label{app:english-only} Additionally, we also evaluate models on 11 English-only natural language inference and commonsense reasoning benchmarks: ARC-C and ARC-E \citep{chollet2019measureintelligence}, BoolQ \citep{clark2019boolq}, CommonsenseQA \citep{talmor-etal-2019-commonsenseqa}, Hellaswag \citep{zellers-etal-2019-hellaswag}, MMLU \citep{hendrycks2021mmlu}, OpenBookQA \citep{OpenBookQA2018}, PIQA \citep{Bisk2020}, SIQA \citep{sap2019siqa}, TruthfulQA \citep{lin-etal-2022-truthfulqa}, and WinoGrande \citep{sakaguchi2019winogrande}. 

We include task-specific evaluations (both multilingual and English-only) to understand the relative merit of different design choices. Typically pretrained models do not perform well at downstream tasks at this point in training, as the models have not yet been optimized for instruction following \citep{wang2022language,ustun-etal-2024-aya,aakanksha2024mixdatamergemodels}, or aligned using reinforcement learning \citep{ahmadian-etal-2024-back,dang2024ayaexpansecombiningresearch}. Hence, we do not expect state-of-the-art performance, but rather evaluate the relative signal of different variants.

\begin{table}[t]
\centering
\resizebox{0.58\textwidth}{!}{ 
\begin{tabular}{ll|ccc}
    \toprule
    Cluster & Tokenizer & Belebele & M-MMLU & EN Tasks \\
    \cmidrule(lr){3-5} 
        & & \multicolumn{3}{c}{\textsc{\textbf{Primary Languages}}} \\
    \midrule
    \multirow{2}{*}{European} & \textsc{Cluster} & 41.4 & 31.1 & 48.5 \\
                              & \textsc{Universal} & 41.9 & 30.9 & 48.4 \\
    \midrule
    \multirow{2}{*}{Asian} & \textsc{Cluster} & 38.2 & 29.6 & 48.2 \\
                           & \textsc{Universal} & 38.1 & 28.9 & 48.1 \\
    \midrule
    Middle East & \textsc{Cluster} & 38.1 & 29.2 & 49.1 \\
    \& Indic    & \textsc{Universal} & 36.5 & 28.6 & 48.2 \\
    \bottomrule
\end{tabular}%
}
\caption{Comparison of \textsc{Cluster} vs. \textsc{Universal} tokenizers during the \textbf{pretraining} on the primary languages across three regional clusters. Performance with \textsc{Universal} tokenizer is comparable to \textsc{Cluster} tokenizers across all the geo-cluster models. 
}
\label{tab:pretraining-results}
\end{table}

\begin{table}[t]
\centering
\resizebox{\textwidth}{!}{
\begin{tabular}{l|*{14}{c}}
    \toprule
     & bul & cat & ces & dan & deu & ell & est & eus & fin & fra & hrv & hun & ita & lit \\
    \midrule
    \textsc{Uniform}   & 41.0 & 43.7 & 42.7 & 41.5 & 43.6 & 41.2 & 35.7 & 38.4 & 34.6 & 45.0 & 42.1 & 36.6 & 40.6 & 41.0 \\
    \textsc{Universal} & 42.1 & 46.5 & 45.5 & 41.1 & 44.3 & 42.7 & 37.9 & 40.6 & 32.4 & 45.3 & 42.6 & 37.9 & 40.1 & 43.1 \\
    & \textcolor{blue}{(+1.1)} & \textcolor{blue}{(+2.8)} & \textcolor{blue}{(+2.8)} & \textcolor{red}{(–0.4)} & \textcolor{blue}{(+0.7)} & \textcolor{blue}{(+1.5)} & \textcolor{blue}{(+2.2)} & \textcolor{blue}{(+2.2)} & \textcolor{red}{(–2.2)} & \textcolor{blue}{(+0.3)} & \textcolor{blue}{(+0.5)} & \textcolor{blue}{(+1.3)} & \textcolor{red}{(–0.5)} & \textcolor{blue}{(+2.1)} \\
    \midrule
    \midrule
     & lvs & nld & nob & pol & por & ron & rus & slk & slv & spa & srp & swe & ukr & Average \\
    \midrule
    \textsc{Uniform}   & 42.3 & 42.7 & 42.3 & 38.4 & 41.0 & 40.8 & 41.0 & 42.4 & 39.5 & 42.5 & 42.4 & 42.9 & 40.9 & 41.0 \\
    \textsc{Universal} & 40.5 & 42.7 & 42.8 & 39.8 & 43.8 & 41.2 & 41.4 & 43.5 & 41.8 & 43.8 & 43.4 & 43.4 & 40.0 & 41.9 \\
    & \textcolor{red}{(–1.8)} & \textcolor{blue}{(+0.0)} & \textcolor{blue}{(+0.5)} & \textcolor{blue}{(+1.4)} & \textcolor{blue}{(+2.8)} & \textcolor{blue}{(+0.4)} & \textcolor{blue}{(+0.4)} & \textcolor{blue}{(+1.1)} & \textcolor{blue}{(+2.3)} & \textcolor{blue}{(+1.3)} & \textcolor{blue}{(+1.0)} & \textcolor{blue}{(+0.5)} & \textcolor{red}{(–0.9)} & \textcolor{blue}{(+0.9)} \\
    \bottomrule
\end{tabular}
}
\caption{Comparison of \textsc{Universal} vs. \textsc{Uniform} tokenizer performance on Belebele, when used for pretraining of Euro cluster model. In tokenizer training, we use all the languages (67 languages; primary and expanded subsets) and only vary the language weighting. \textsc{Universal} tokenizer with balanced weighting using language buckets outperforms \textsc{Uniform} weighting in 21 European languages out of 27, with a relative gain of 2.2\% (41.9 vs 41.0) on average.}
\label{tab:uniform-vs-universal}
\end{table}
 
\section{Results on Pretraining Performance}
\label{sec:pretraining-results}

\begin{sectionfindings}
    \item Pretraining with \textsc{Universal} tokenizer produces well-rounded models with competitive results on a variety of tasks for the primary languages, differing from the \textsc{Cluster} tokenizer models by no more than 0.5\% average accuracy across tasks and clusters.
    \item Our specialized tokenizer weighting that uses language buckets to balance the data availability, leads to better pretraining performance up to 2.8 accuracy increase as measured in the Euro cluster model.    
\end{sectionfindings}

In this section, we first benchmark the performance of our pretrained models, to ensure that using a \textsc{Universal} tokenizer doesn't cause degradations in performance, as may be expected when using a tokenizer optimized for a broader set of languages than the primary set.

\textbf{\textsc{Universal} tokenizer does not compromise performance on primary languages.} As seen in Table \ref{tab:pretraining-results}, we find that our expanded \textsc{Universal} tokenizer is remarkably competitive against \textsc{Cluster} across the geo-cluster models. The difference in pretraining performance is less than at most 1\% average accuracy in English tasks. The highest performance difference between \textsc{Universal} and \textsc{Cluster} tokenizers for multilingual tasks is only 1.6\% average accuracy on Belebele for ME-Indic cluster (38.1\% vs 36.5 \%). Overall, we observe minimal trade-offs in performance on primary cluster languages switching to \textsc{Universal} tokenizer.

In fact, we observe that the \textsc{Universal} tokenizer leads to a slight increase on average on Belebele for Euro cluster (41.9 vs 41.4) and achieves much closer performance for Asian cluster (38.1 vs 38.2). As additional validation, Figure \ref{fig:belebele-pretraining} in Appendix \ref{app:pretraining} shows the progression of average Belebele performance for both tokenizers for Euro cluster models during pretraining. The \textsc{Universal} tokenizer achieves approximately similar performance throughout the whole pretraining, also suggesting the same trend in a longer pretraining run. Overall, these results showcase that using a \textsc{Universal} tokenizer doesn't spell any significant performance degradation in pretraining for the primary languages. 

\textbf{Balanced language weighting with language buckets for tokenizer training leads to better pretraining performance.} As described in Section \ref{sec:tokenizer-training} on tokenizer training, we weight the languages using buckets formed by script and language family, balanced against data availability. In order to motivate this weighting scheme, we compare pretraining performance of \textsc{Universal} tokenizer against a baseline tokenizer (\textsc{Uniform}), where all languages are uniformly weighted except English.\footnote{For both tokenizers' weighting, we use a fixed proportion of 30\% for English due to much larger data volume and much higher diversity in available data. This also ensures a fair comparison between tokenizer weighting.} We conduct this ablation in the Euro cluster, where the number of primary languages is the highest. In tokenizer training, we use all the languages (62 languages; primary and expanded subsets) and only vary the language weighting. As shown in Table \ref{tab:uniform-vs-universal}, \textsc{Universal} tokenizer with balanced weighting using language buckets outperforms \textsc{Uniform} weighting in 21 European languages out of 27, with a relative gain of 2.2\% (41.9 vs 41.0) on average. Further validating pretraining results, we provide the comparison for compression performance between these two tokenizers in Appendix \ref{app:compression-ratio}, where the results show better overall compression in \textsc{Universal} tokenizer.

\begin{table}[t]
\centering
\resizebox{0.55\textwidth}{!}{ 
\begin{tabular}{ll|cl}
    \toprule
    Cluster & Tokenizer & \multicolumn{2}{c}{Dolly Win Rates (\%)}\\
    \cmidrule(lr){3-4} 
     & &  \textsc{\textbf{Primary}} & \textsc{\textbf{Expanded}} \\
    \midrule
    \multirow{2}{*}{European} & \textsc{Cluster} & 42.8 & ~~~17.6   \\
                              & \textsc{Universal} & 42.8 & ~~~37.4 ~(\textcolor{blue}{+19.9}) \\
    \midrule
    \multirow{2}{*}{Asian} & \textsc{Cluster} & 41.7 & ~~~11.7 \\
                           & \textsc{Universal} & 41.8 & ~~~29.5 ~(\textcolor{blue}{+17.8}) \\
    \midrule
    Middle East & \textsc{Cluster} & 39.7 & ~~~22.8 \\
    \& Indic              & \textsc{Universal} & 40.2 & ~~~41.8 ~(\textcolor{blue}{+18.9})\\
    \bottomrule
\end{tabular}%
}
\caption{Win rates after \textbf{continued pretraining} on primary and expanded language subsets. The \textsc{Universal} tokenizer matches \textsc{Cluster} performance on primary languages and shows large gains (up to 19.9\%) on average of expanded language subsets across all clusters.}
\label{tab:continued-pretraining}
\end{table}

\begin{table}[t]
\centering
\resizebox{0.50\textwidth}{!}{ 
\begin{tabular}{l|cc}
    \toprule
     & \textsc{Cluster} & \textsc{Universal} \\
    \midrule
    European & 27.2 & 37.4 ~(\textcolor{blue}{+10.2}) \\
    Asian & 18.8 & 34.3 ~(\textcolor{blue}{+15.7}) \\
    Middle East \& Indic & 23.31  & 41.1 ~(\textcolor{blue}{+17.8}) \\
    \bottomrule
\end{tabular}%
}
\caption{Win rates on expanded languages after \textbf{targeted adaptation}. The \textsc{Universal} tokenizer shows better performance (up to 17.8\%) across all clusters over the baseline \textsc{Cluster} tokenizer.}
\label{tab:sft-expanded-win-rates}
\end{table}

\newpage
\section{Results on Enhanced Multilingual Plasticity} 

\subsection{Benefits of Plasticity in Continued Pretraining}

\begin{sectionfindings}
    \item The \textsc{Universal} tokenizer leads to significantly higher \textit{plasticity} over the \textsc{Cluster} specific tokenizers in continued pretraining (on primary \& expanded languages) with an average increase in win rate of 18.9\% for the expanded language subsets across clusters. 
    \item The \textsc{Universal} tokenizer enables adaptation boost with no drop in primary languages compared to \textsc{Cluster} tokenizer, with a near identical performance (with a slight increase of 0.3\% on average) on downstream open-ended generations.   
\end{sectionfindings}

In this section, we ask: \textit{Does varying the approach for the tokenizer lead to plasticity benefits after continued pretraining on both primary and expanded languages?} 

\noindent\textbf{Models trained with the \textsc{Universal} tokenizer demonstrate significantly higher win rates on the \textsc{expanded subset}.} Figure \ref{fig:all-langs-cooldown} and Table \ref{tab:continued-pretraining} show results of evaluation across the Euro, Asian, and ME-Indic clusters, with 5 languages belonging to each \textsc{primary language subset} and 10 in the \textsc{expanded subset} belonging to the other two clusters. We see that the \textsc{Universal} tokenizer achieves an average gain of 18.9\% in win rates across all three geo-cluster models on the expanded subsets, compared to models trained with the \textsc{Cluster} tokenizer. Improvement is consistent across clusters, where we find +19.9\%, +17.8\%, and +18.9\% increase in win rate for Euro, Asian, and ME-Indic cluster models, respectively. Among all the expanded languages, Persian (+25.8\%), Hindi (+23.3\%), and Vietnamese (+22.0\%) show the highest benefit from the \textsc{Universal} tokenizer in the Euro, Asian, and ME-Indic clusters, respectively.  

\noindent\textbf{\textsc{Universal} tokenizer preserves performance on multiple clusters.} While the \textsc{Universal} tokenizer provides significant gains on the expanded languages, in Figure \ref{fig:all-langs-cooldown}, we observe that the performance on primary languages is nearly the same across both tokenizers in all three clusters. 
There is only a 0.3\% win rate difference across all clusters for primary languages when comparing tokenizers, where \textsc{Universal} tokenizer even leads to a slight increase over the \textsc{Cluster} tokenizer in the European, Asian, and ME-Indic models by 0.3\%, 0.1\%, and 0.5\%, respectively. This is beneficial, as it suggests no trade-offs of improving plasticity for an expanded set of languages by using the \textsc{Universal} tokenizer for the primary languages a provider is interested in when developing a model. 

\begin{figure*}[t]
\centering
    \begin{subfigure}[t]{0.58\textwidth} 
    \centering
        \includegraphics[width=\textwidth]{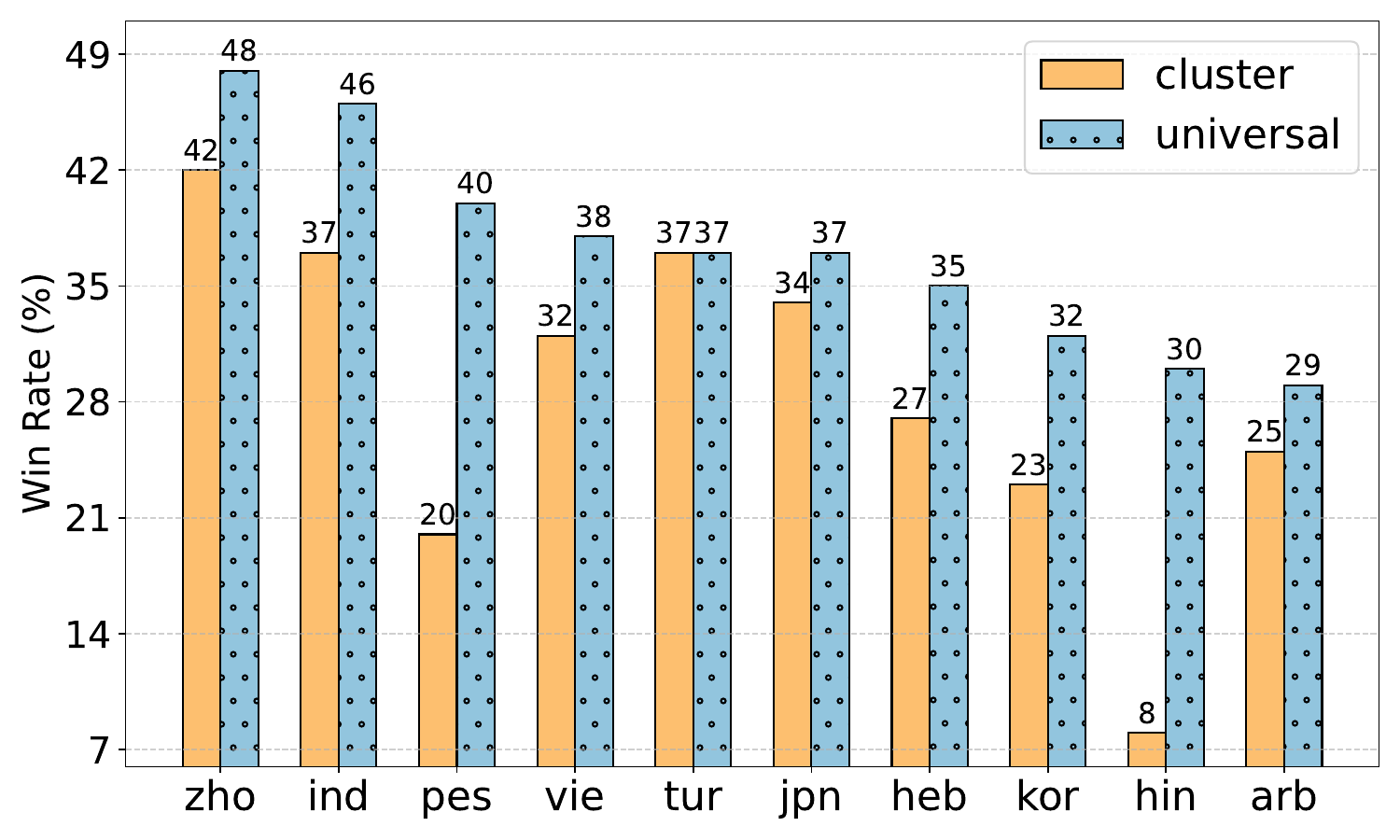}
        \caption{\textbf{Expanded} languages}
    \label{fig:sft-euro}
    \end{subfigure}
    \begin{subfigure}[t]{0.405\textwidth} 
    \centering
        \includegraphics[width=\textwidth]{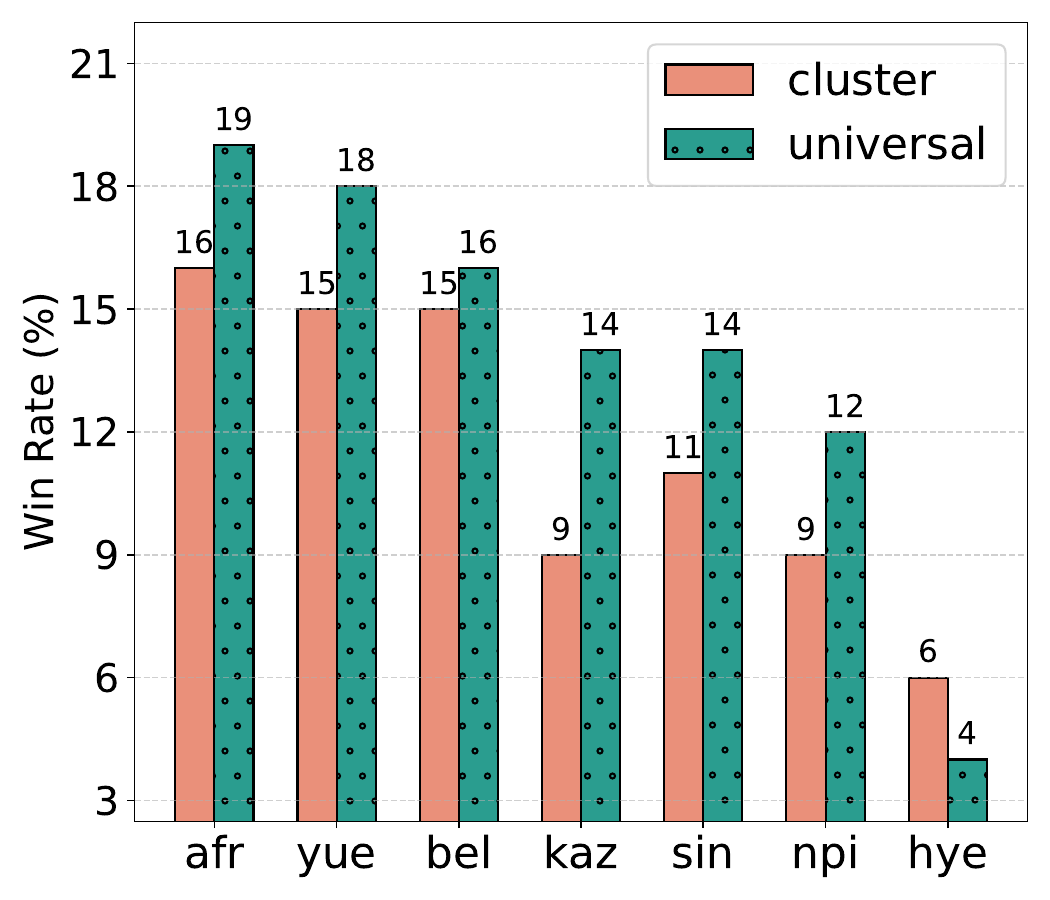}
        \caption{Fully \textbf{unseen} languages.}
    \label{fig:unseen-langs}
    \end{subfigure}
    \caption{
        Language-specific results after \textbf{targeted adaptation} through SFT for the Euro cluster model to \textbf{expanded} languages (a) and fully \textbf{unseen} languages (b). \textsc{Universal} tokenizer outperforms the \textsc{Cluster} tokenizer in both language subsets, where the relative gains over the cluster-specific tokenizer go up to 22\% on expanded languages (Hindi), and 5\% on unseen languages (Kazakh).  
    }
    \label{fig:post-training}
\end{figure*} 

\subsection{Benefits of Plasticity in Targeted Adaptation} 

\begin{sectionfindings}
    \item For the targeted language adaptation through SFT, where only the expanded language data is introduced, the \textsc{Universal} tokenizer is much more performant than the \textsc{Cluster} tokenizer with 14.6\% average increase across languages and geo-clusters.
    \item \textsc{Universal} tokenizer enables multilingual plasticity not only for languages seen during tokenizer training but also for fully \textsc{unseen} languages with an average improvement of 2\% on 7 under-resourced languages over the \textsc{Cluster} tokenizer. 
\end{sectionfindings}

An experimental setting of great interest is the more realistic scenario where a downstream developer only has access to data in the \textsc{expanded} languages. To mimic this scenario, we evaluate the impact of our interventions when only supervised fine-tuning only on the expanded language subset is feasible. 

\textbf{\textsc{Universal} tokenizer outperforms \textsc{Cluster} tokenizers by high margins in targeted adaptation for expanded language set.} Table \ref{tab:sft-expanded-win-rates} shows average win rates of \textsc{Universal} and \textsc{Cluster} tokenizers for each geo-cluster. \textsc{Universal} tokenizer achieves 10.2\%, 15.7\%, and 17.8\% relative win rate gains over the  \textsc{Cluster}-specific tokenizers for Euro, Asian, and ME-Indic clusters, respectively. In Figure \ref{fig:sft-euro}, we also plot individual language gains for the Euro cluster. \textsc{Universal} consistently enables higher plasticity than \textsc{Cluster} tokenizer where the relative gains go up to 22.0\% and 20.0\% in Hindi and Farsi, respectively. 

\textbf{\textsc{Universal} tokenizer also provides large gains in targeted adaptation for fully unseen language set.} In the most extreme setting, we evaluate the benefits of our tokenizer intervention for adaptation to languages that are \textbf{fully unseen} in both tokenizer and pretraining. Figure \ref{fig:unseen-langs} shows results on supervised fine-tuning experiments on 7 unseen languages.\footnote{Afrikaans, Kazakh, Belarusian, Cantonese, Nepali, Armenian, Sinhala} It is critical that all these languages are extremely under-resourced, and this adaptation is performed in a low-data environment, as this is representative of the constraints faced by developers in these languages.

\textbf{In the most extreme setting, \textsc{Universal} tokenizer enables 
 unseen language gains.} We find that \textsc{Universal} tokenizer enables improvements over the \textsc{Cluster} tokenizer on the unseen languages with an average gain of 2.0\% in win rates, where it goes up to 5.0\% in Nepali. Given that the downstream performance is generally lower for these languages due to their absence in the tokenizer and pretraining, and also constraints on available data (only 15k per language), we find this to be a promising direction of future research and another reason to invest in more flexible tokenizer design. 

\newpage

\begin{figure*}[t]
\centering
    \begin{subfigure}[t]{0.45\textwidth} 
    \centering
        \includegraphics[width=\textwidth]{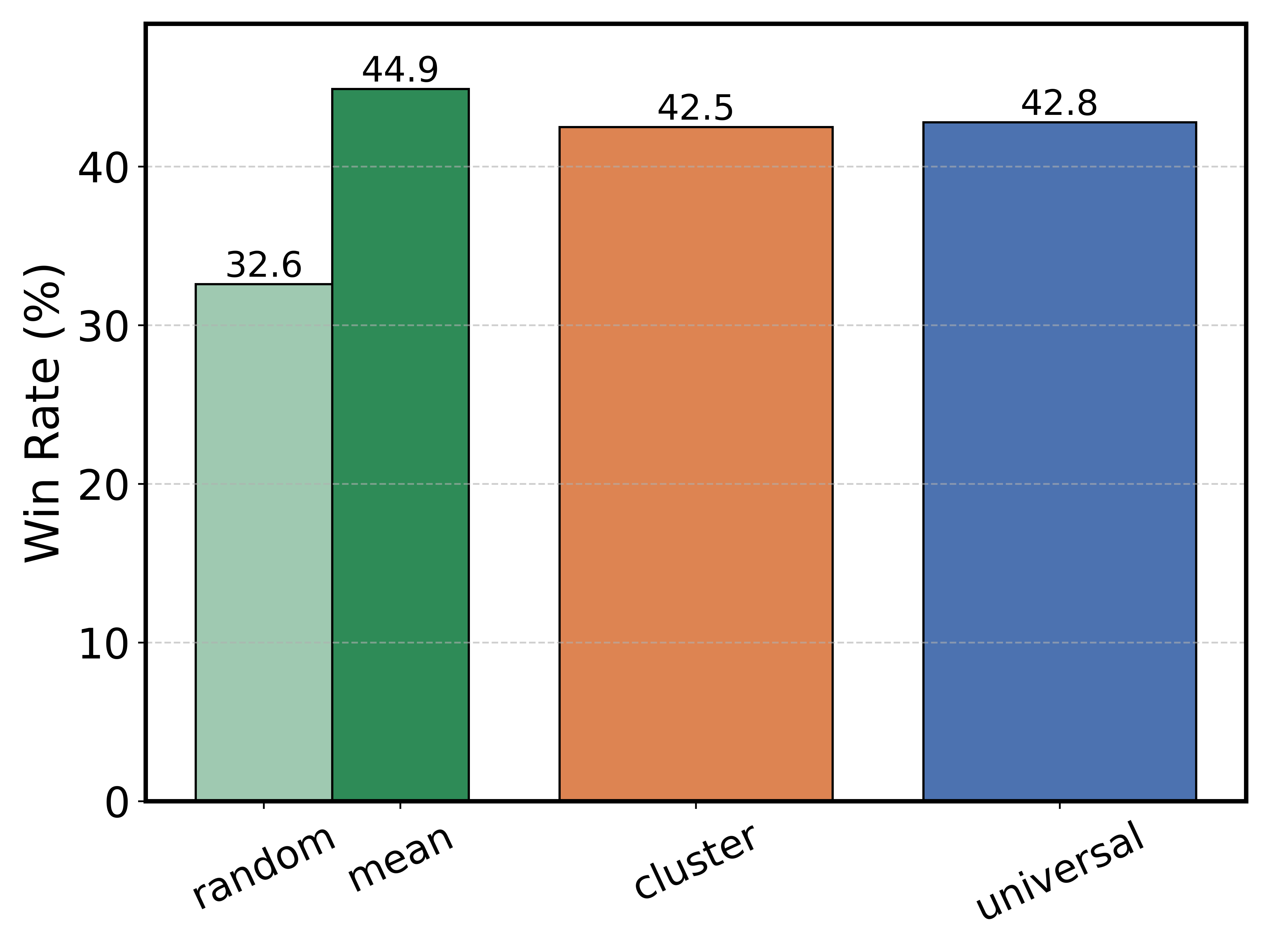}
        \caption{\textbf{Primary} languages.}
        \label{fig:cva-id}
    \end{subfigure}
    \hspace{0.2cm}
    \begin{subfigure}[t]{0.45\textwidth} 
    \centering
        \includegraphics[width=\textwidth]{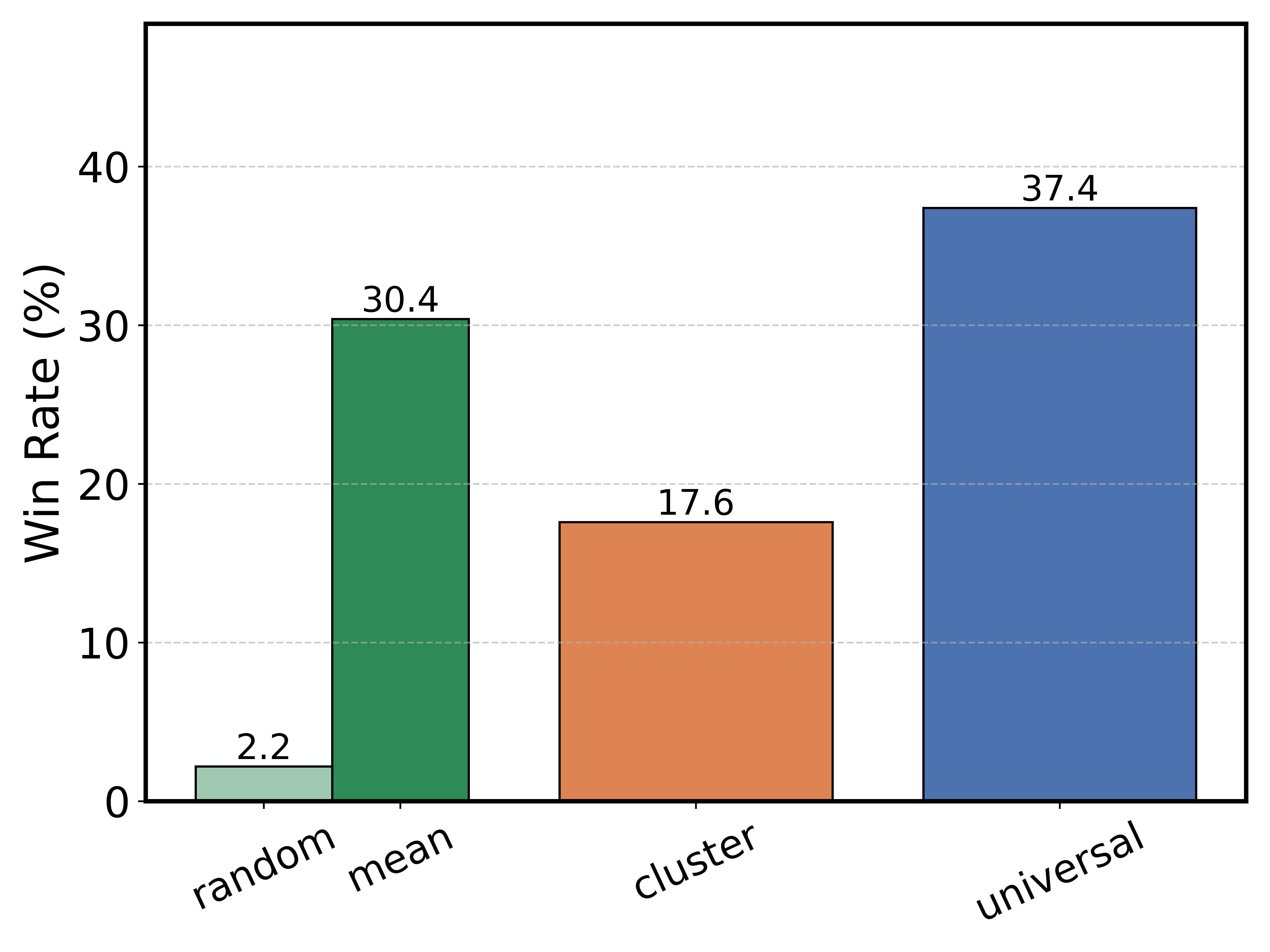}
        \caption{\textbf{Expanded} languages}
        \label{fig:cva-ood}
    \end{subfigure}
    \caption{
        Win rates after continued pretraining, comparing Cross-lingual Vocabulary Adaptation (CVA) through tokenizer replacement. After the pretraining, the \textsc{Cluster} tokenizer is replaced with the \textsc{Universal} tokenizer, the embeddings for the shared tokens are preserved and the new tokens are either randomly initialized (\texttt{random}) or by the average of shared embeddings (\texttt{mean}). The \textsc{Universal} tokenizer significantly outcompetes both CVA approaches on expanded languages.
        }
    \label{fig:cva-results}
\end{figure*}

\section{Key Discussions}

\subsection{Comparison with cross-lingual vocabulary adaptation}

\begin{sectionfindings}
\item \textsc{Universal} tokenizer outperforms cross-lingual vocabulary adaptation (CVA) by 7\% in the expanded languages where the vocabulary is replaced (before the continued pretraining), and the new tokens are initialized by average embeddings of the shared tokens. 
\item CVA through tokenizer replacement fails to achieve competitive performance even with the \textsc{Cluster} tokenizer when the new tokens are initialized randomly. CVA (\texttt{random}) only reaches 2.2\% win rate in the expanded languages and falls behind the \textsc{Universal} tokenizer by 35.2\% relative drop in win rate.
\end{sectionfindings}

Cross-lingual vocabulary adaptation (CVA) \citep{yamaguchi2024effectivelyexpandvocabularyllms} aims to adapt the existing tokenizer and hence the token embeddings to the new languages through expansion or replacement after pretraining, and is a common approach for language adaptation. A more detailed overview of CVA can be found in Section \ref{sec:related-works}. In this ablation, we ask: \textit{how does the \textsc{Universal} tokenizer compare with CVA for adapting new languages?}

To have a fully comparable setup with our body of experiments, we took the pretrained Euro cluster model that is trained with the \textsc{Cluster} tokenizer and replaced the tokenizer with the \textsc{Universal} tokenizer. Token embeddings that are shared between \textsc{Cluster} and \textsc{Universal} tokenizers are preserved, and new tokens are either randomly initialized by sampling from a normal distribution (\texttt{random}), or with the average of the shared embeddings (\texttt{mean}). After vocabulary replacement and token initialization, we follow the same continued pretraining described in Section \ref{sec:strategies}. 

The results for this ablation are given in Figure \ref{fig:cva-results}. We find that when randomly initializing the new tokens, CVA (tokenizer replacement) fails to achieve comparable performance even against the \textsc{Cluster} tokenizer, and significantly lags behind the \textsc{Universal} tokenizer by 15.4\% and 35.2\% win rates for primary and expanded languages respectively. Notably, initializing the new tokens with the mean of the shared vocabulary (\texttt{mean} in Figure \ref{fig:cva-results}), outperforms random initialization. While tokenizer replacement (\texttt{mean}) is an improvement over the unadapted \textsc{Cluster} tokenizer by 12.8\% relative increase in win rates for expanded languages, our \textsc{Universal} tokenizer leads to better adaptation performance by 7\% difference in average win rate (37.4\% vs 30.4\%). Interestingly, CVA (\texttt{mean}) achieves slightly higher performance in the primary languages by 2.1\% average win rate. Overall, these results show that it is more effective to use a \textsc{Universal} tokenizer from the start, rather than substituting it in after pretraining.

\subsection{Adaptation Efficiency with the Universal Tokenizer}

\begin{sectionfindings}
\item \textsc{Universal} tokenizer enables +8x faster adaptation in terms of sample efficiency and +2x higher performance for downstream adaptation compared to \textsc{Cluster} tokenizer. 
\end{sectionfindings}

In this ablation, we evaluate how much faster adaptation takes place with the \textsc{Universal} tokenizer. Faster adaptation means much fewer resources necessary, namely cost, which is of great interest to practitioners who want to adapt an LLM to expanded language coverage. 

To evaluate the adaptation speed, in comparison to \textsc{Cluster} tokenizer, we evaluated the intermediate checkpoints on the expanded languages during continued pretraining for Euro cluster models. Figure \ref{fig:adaptation-speed} shows average win rates on 10 expanded languages. As seen in the plot, in only 300 steps \textsc{Universal} tokenizer reaches the level of performance that \textsc{Cluster} achieves at 2500 steps, showing +8x faster adaptation. Given that 300 steps correspond to nearly 150K samples (compared to 1.3M at 2500 steps), \textsc{Universal} tokenizer requires also much less data to achieve the same performance with the end performance of the baseline, confirming the effectiveness of our proposal. 

\begin{figure*}[t]
\centering
    \begin{subfigure}[t]{0.335\textwidth} 
    \centering
        \includegraphics[width=\textwidth]{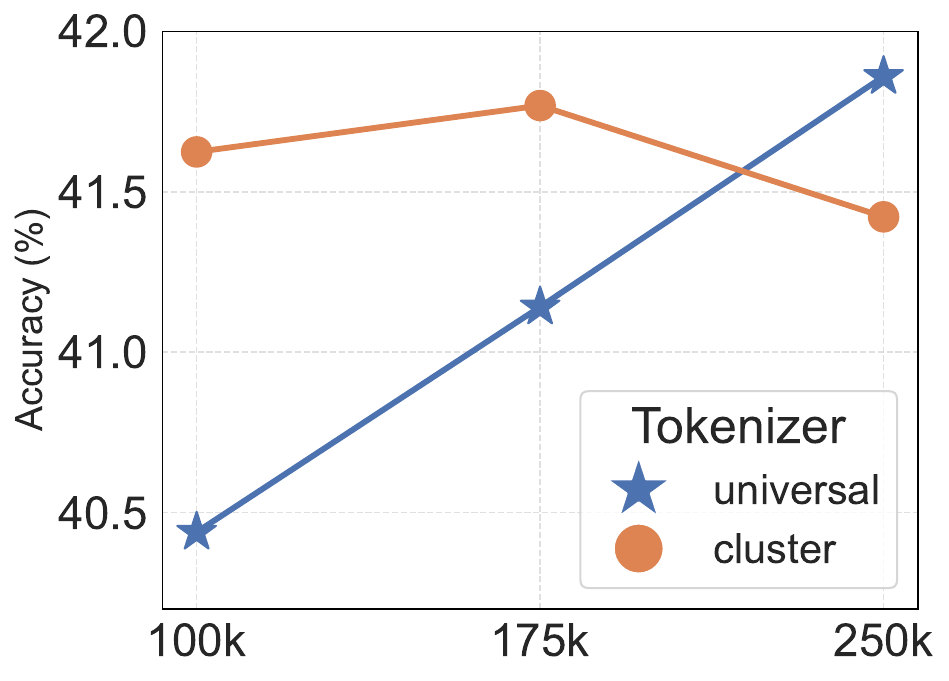}
        \caption{Effect of vocabulary size.}
        \label{fig:vocab-size}
    \end{subfigure}
    \hspace{0.2cm}
    \begin{subfigure}[t]{0.63\textwidth} 
    \centering
        \includegraphics[width=\textwidth]{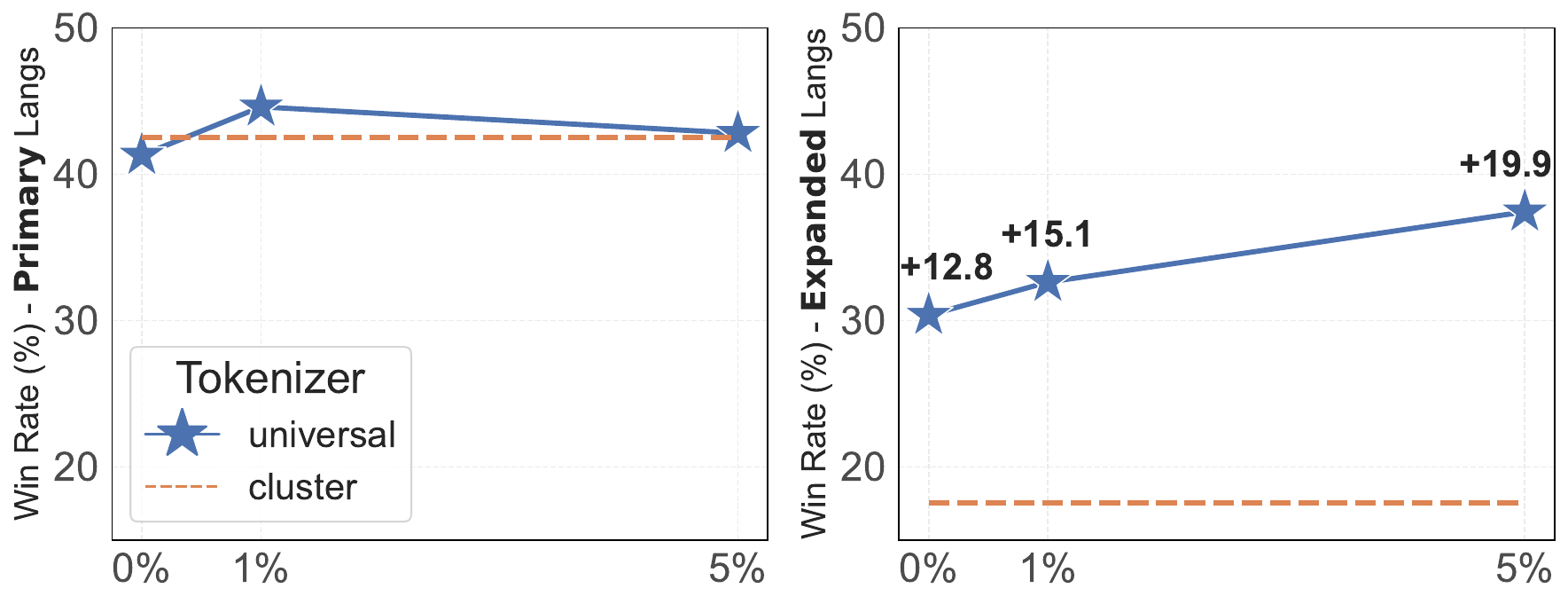}
        \caption{Effect of the expanded language data in pretraining}
        \label{fig:multilingual-percentage}
    \end{subfigure}
    \caption{
        (a) \textsc{Universal} tokenizer requires a larger vocabulary size to achieve the same (or better) pretraining performance with the \textsc{Cluster} tokenizers on \textbf{primary} languages, as evaluated in Belebele. 
        (b) \textsc{Universal} tokenizer exhibits significantly higher adaptation gains over the \textsc{Cluster} tokenizer even when there is no data from the expanded languages added in pretraining.}
    \label{fig:tokenizer-scaling}
\end{figure*}

\subsection{Necessity of Large Vocabulary Size}
\label{sec:vocab-size}

\begin{sectionfindings}
\item The \textsc{Universal} tokenizer demonstrates effectiveness without compromising pretraining performance for primary languages, but it necessitates a large vocabulary size, such as 250,000 subwords, to achieve optimal results.
\end{sectionfindings}

In the previous sections, we establish greater performance in multilingual plasticity with the \textsc{Universal} compared to \textsc{Cluster} tokenizer. In this ablation, we ask: \textit{What is the required vocabulary size for the \textsc{Universal} tokenizer to avoid performance degradation on primary pretraining languages?} 

To determine the optimal vocabulary size, we run additional pretraining experiments where we vary the vocabulary size from 100,000 tokens to 250,000 tokens while adjusting the model parameters so that the total number of trainable parameters remains constant. We evaluate the performance for the primary pretraining languages on Belebele. Results are shown in Figure \ref{fig:vocab-size}. The models trained with \textsc{Cluster} tokenizers don't vary much in performance, and surpass the \textsc{Universal} tokenizer at small vocabulary sizes (100k and 175k). However, the \textsc{Universal} tokenizer scales performance with the vocabulary size, and overtakes the \textsc{Cluster} tokenizer at 250k vocabulary size. Our findings are consistent with previous work that shows the benefits of large vocabularies \citep{tao2024, huang2025overtokenized}, and suggests investment in universal tokenizers require a reallocation of weights to ensure a proper vocabulary budget. Based on this ablation, we use the vocabulary size of 250k in our main pretraining runs.

\subsection{Presence of Expanded Language Subset in Pretraining} 
\label{sec:expanded-subset-percentage}

\begin{sectionfindings}
\item The \textsc{Universal} tokenizer achieves a performance boost of 12.8\% over the \textsc{Cluster} tokenizer for the expanded languages even there is no pretraining data is used for these languages. However, including a minimal data up to 5\% increases adaptation performance on the expanded languages from 12.8\% to 19.8\% win rates, without hurting performance in primary pretraining languages.
\end{sectionfindings}

In the large and often noisy datasets used to train LLMs, there is often language contamination \citep{blevins-zettlemoyer-2022-language}. Therefore, it can be difficult to claim a language is truly ``new''. In our final ablation, to test the robustness of our claims of plasticity under different assumptions of multilingual data presence, we evaluate 0\%, 1\%, and 5\% proportion of expanded languages in pretraining for the European cluster. 

Figure \ref{fig:multilingual-percentage} shows that even in the most conservative case of 0\% multilingual percentage for the new languages (the expanded subset), the \textsc{Universal} tokenizer exhibits 12.8\% gain in win rate as compared to the \textsc{Cluster} tokenizer. Notably, increasing this percentage up to 5\% does not hurt performance in primary pretraining languages, but increases adaptation performance on the expanded languages from 12.8\% to 19.8\% win rates.

\newpage
\section{Related Work}
\subsection{Multilingual Tokenizers}
Tokenization remains a key challenge for multilingual models, particularly due to inefficiencies and disparities across scripts.  
\citet{petrov2023} show that many standard or English-centric tokenizers disproportionately fragment non-Latin scripts, sometimes producing up to 15 times more tokens than for equivalent English content.
These imbalances can lead to practical consequences such as increased API usage costs \citep{ahia-etal-2023-languages, petrov2023}, longer inference times \citep{hofmann-etal-2022-embarrassingly, sun-etal-2023-multi}, and reduced usable context window for non-English languages \citep{velayuthan-sarveswaran-2025-egalitarian, ahia-etal-2023-languages}, and degraded downstream task performance \citep{goldman-etal-2024-unpacking, gow-smith-etal-2022-improving, fujii-etal-2023-different}. 
Moreover, inefficient tokenization for non-English languages can inflate training costs by up to 68\% \citep{ali-etal-2024-tokenizer}. 

To support multilinguality, the mT5 model introduced a 250k SentencePiece vocabulary with byte-fallback, pretrained on 101 languages using temperature-based sampling to balance high- and low-resource languages \citep{xue-etal-2021-mt5}.
Additionally, recent work has explored tokenization approaches tailored to better reflect the structure of diverse writing systems. 
For example, Grapheme-aware tokenization, using Unicode grapheme clusters as atomic units via methods like Grapheme Pair Encoding (GPE) \citep{velayuthan-sarveswaran-2025-egalitarian} or MYTE, a segmentation strategy motivated by morphemes \citep{Limisiewicz_2024}, offers improved representation for complex scripts.

\subsection{Language Adaptation Post-Training}
\label{sec:related-works}
There are a number of ways that language adaption of pretrained language models (PLMs) is commonly approached. In terms of additional training, continued pretraining (CPT) and supervised fine-tuning (SFT) are the most common methods. Continued pretraining involves extended training on the adaptation language corpora \citep{han-eisenstein-2019-unsupervised, muller-etal-2021-unseen}, but requires a significant amount of data, which may not be accessible for lower-resourced languages. Supervised fine-tuning is also a standard approach \citep{kumar-etal-2022-indicnlg, masakhaner, cahyawijaya-etal-2021-indonlg}, requiring less data than CPT, but possibly leading to catastrophic forgetting of capabilities from pretraining \citep{NEURIPS2019_fa7cdfad, chaudhry2019tinyepisodicmemoriescontinual}. In particular, instruction fine-tuning is popular in order to impart instruction-following capabilities as well \citep{gala2024airavataintroducinghindiinstructiontuned}. Claims as to the advantages of one over the other are mixed- \cite{ebrahimi-kann-2021-adapt} find that in their setup, CPT is more effective than SFT, and \citet{yong-etal-2023-bloom} find the opposite. 

A primary challenge is unsupported scripts and languages in the tokenizer. Cross-lingual vocabulary adaptation (CVA) modifies the existing tokenizer to accommodate additional languages \citep{yamaguchi-etal-2024-empirical}, and requires continued pretraining in the target language to sufficiently adapt \citep{Fujii2024ContinualPF}. There are two common approaches to CVA -- vocabulary expansion, where new tokens are added from the target and shared tokens are reused \citep{wang-etal-2020-extending, pfeiffer-etal-2021-unks}, or vocabulary replacement, where the vocabulary is entirely replaced. Embeddings corresponding to new tokens may be initialized randomly, using heuristics such as the average of some corresponding tokens in the original vocabulary \citep{minixhofer-etal-2022-wechsel, dobler-de-melo-2023-focus, downey-etal-2023-embedding}, or based on auxiliary models \citep{ostendorff2023efficientlanguagemodeltraining}. Switching out the tokenizer is cumbersome; one possible method is by training a hypernetwork that maps vocab of the new tokenizer to existing embeddings \citep{minixhofer2025}. This method requires continued training to close the performance gap, but even then doesn't surpass it. It has also been proposed to transliterate languages to Latin script to circumvent unsupported scripts \citep{muller-etal-2021-unseen}, but this approach is limited by transliteration performance.

It's also possible to elicit competitive performance in unseen languages without additional training or fine-tuning steps. \citet{tanzer2024benchmarklearningtranslatenew} leverage long context lengths of SoTA language models to unlock translation capabilities in an extremely low-resource, entirely unseen language, including context from a grammar book on the language in the prompt. \citet{cahyawijaya-etal-2024-llms} use in-context learning cross-lingually, incorporating task examples from different higher-resourced languages in the prompt to enable transfer of task-specific capabilities to a lower-resourced language.

\section*{Conclusion}
In this work, we explore what cheap interventions in pretraining can increase plasticity in downstream optimization stages. We conduct an extensive study involving different tokenization strategies, three language adaptation strategies involving different assumptions about data access across 70 different languages. We find that in all cases, a model trained using a \textsc{universal} tokenizer with broad language coverage is able to adapt to languages outside of the primary pretraining set far better, with average win rate improvements up to 20.2\% in continued pretraining and 17.8\% in targeted adaptation to expanded languages. Even in the challenging low-data setting of completely unseen languages, the \textsc{universal} tokenizer shows gains up to 5\%. At the same time, there is negligible performance impact to the primary pretraining languages. Investing in a massively multilingual tokenizer up-front pays off in language adaptation down the line.

\section{Limitations}

\noindent\textbf{Language coverage.} Our experiments involve 69 languages covering a diverse range of languages and scripts, where we systematically investigate the impact of multilingual data on tokenizer training. Although we use a comprehensive list of languages, there are many more languages in the world, which requires attention of the research community. We hope our work encourages even broader language coverage in state-of-the-art language models.
    
\noindent\textbf{Tokenization algorithm.} In this work, we focused only on the BPE algorithm for tokenizer training, which is the most widely used method for language models. This choice was dictated by the high computational cost of each ablation, which required significant compute resources. However, we believe our findings on multilingual coverage would apply to the other tokenizers, such as Unigram tokenizer \citep{kudo2018subword} or byte or character-level tokenization \citep{xue-etal-2022-byt5,clark-etal-2022-canine}. We leave this exploration to future research. 

\noindent\textbf{Model size.} All of our pretraining experiments were conducted on 3.3B parameter models with a 100B token budget, which is already an enormous undertaking for resources and compute costs. Given that our results hold for this scale, we anticipate they would also apply to larger models and token budgets, as supported by previous research \citep{biderman2023pythia,longpre-etal-2024-pretrainers,aryabumi2024code}.

\section{Acknowledgments}
We thank Julia Kreutzer, John Dang, Roman Castagné, Aakanksha, and other colleagues at Cohere and Cohere Labs for their support and thoughtful feedback. We also thank Shayne Longpre for his feedback on the preprint. 

\bibliography{main, anthology}

\appendix

\clearpage
\newpage

\section{Additional Tokenizer Details}
\label{app:tokenizer-details}

We use the GPT-4o (\texttt{gpt4-o200k}) regex for pretokenization in all the tokenizers we trained.\footnote{\url{https://github.com/openai/tiktoken/blob/4560a8896f5fb1d35c6f8fd6eee0399f9a1a27ca/tiktoken_ext/openai_public.py\#L95}} Tokenizer training data is sampled from the pretraining data mixture based on the weights described in Section \ref{sec:tokenizer-training}. 
We use the tokenizers \footnote{\url{https://github.com/huggingface/tokenizers}} library to train all the BPE models. We set the \texttt{min\_frequency} argument to 5 on the BPE trainer to control the minimum frequency to merge pairs and do not use normalizers. Finally, we sample 50GB of data to train all the tokenizers.

\section{Additional Training Details}
\label{app:training-details}

\noindent\textbf{Hyperparameters}
We performed a hyperparameter sweep on learning rate (LR), and used $2 \times 10^{-2}$ as the peak LR for all pretraining experiments. We use a batch size of 512, a sequence length of 8192, and a cosine learning rate scheduler with a warmup of 2500 steps. For language adaptation experiments after pretraining, we use a constant LR of $1 \times 10^{-4}$, corresponding to the end LR of the pretraining stage. 

\section{Additional Results}

\subsection{Additional pretraining results}

\begin{figure}[t]
\centering
    \includegraphics[width=0.48\textwidth]{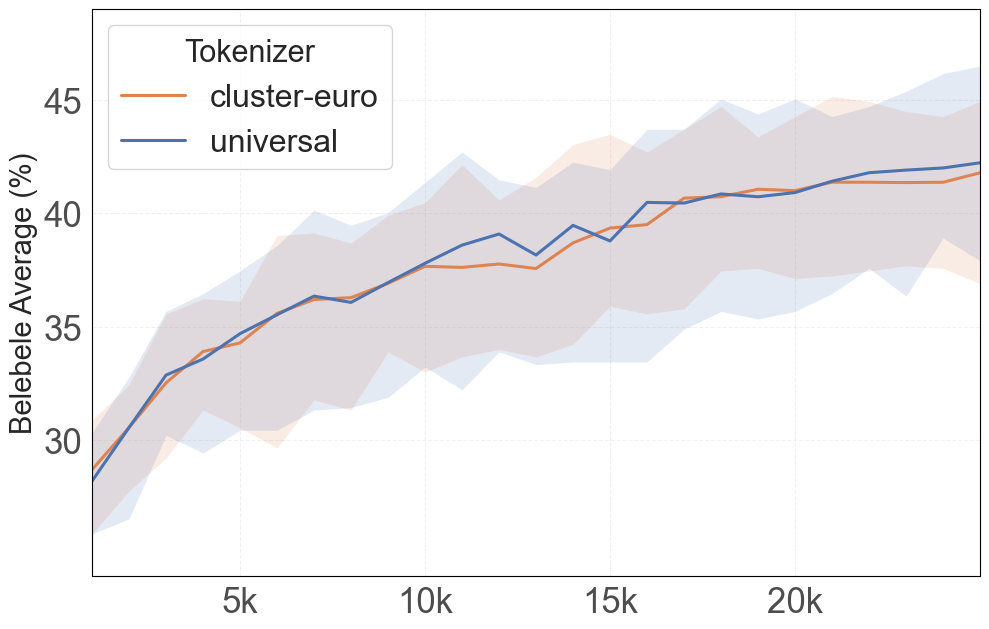}
    \caption{Average performance on primary languages during pretraining (Euro), measured in Belebele. Shaded areas indicate results across languages for each tokenizer. \textsc{Universal} tokenizer shows nearly the same performance with \textsc{Cluster} tokenizer throughout the training, suggesting a similar performance also in much longer pretraining runs.}
    \label{fig:belebele-pretraining}
\end{figure}

Figure \ref{fig:belebele-pretraining} shows pretraining results on the primary language subset (Euro cluster), measured in Belebele, throughout the pretraining run.

\label{app:pretraining}

\subsection{Expanded language adaptation results}

\subsubsection{Continued pretraining}
\label{app:full-cpt-results}

\begin{table}[t]
\centering
\resizebox{0.6\textwidth}{!}{ 
\begin{tabular}{ l | r  c c | r  c c}
    \toprule
     & \multicolumn{3}{c|}{Asian} & \multicolumn{3}{c}{ME-Indic} \\
     & &  \textsc{Cluster} & \textsc{Universal} & & \textsc{Cluster} & \textsc{Universal} \\
    \midrule
     \multirow{5}{*}{\textsc{Primary}} 
     & ind & 56.5 & 46.5 & arb & 37.5 & 43.0 \\
     & jpn & 52.0 & 40.5 & heb & 38.0 & 39.0 \\
     & kor & 46.0 & 40.0 & hin & 41.5 & 38.0 \\
     & vie & 49.5 & 39.5 & pes & 46.0 & 42.5 \\
     & zho & 52.5 & 45.0 & tur & 35.5 & 38.5 \\
     \midrule
     \multirow{10}{*}{\textsc{Expanded}} 
     & arb & 16.0 & 27.0 & deu & 18.0 & 35.5 \\
     & deu & 25.5 & 27.5 & ell & 10.6 & 29.5 \\
     & ell & 9.0 & 29.0 & fra & 28.0 & 43.0 \\
     & fra & 31.5 & 35.5 & ind & 24.5 & 46.5 \\
     & heb & 6.1 & 24.5 & jpn & 20.5 & 46.0 \\
     & hin & 5.1 & 27.5 & kor & 19.0 & 37.5 \\
     & pes & 13.5 & 33.5 & rus & 20.5 & 41.0 \\
     & rus & 17.5 & 38.0 & spa & 35.5 & 46.5 \\
     & spa & 38.0 & 43.0 & vie & 16.0 & 38.0 \\
     & tur & 13.0 & 27.5 & zho & 35.5 & 54.0 \\
    \bottomrule
\end{tabular}%
}
\caption{Full win rate results by language for language adaptation through continued pretraining.}
\label{tab:cpt-full-results}
\end{table}

Table \ref{tab:cpt-full-results} presents the win rates by language after continued pretraining, divided into \textsc{Primary} and \textsc{Expanded} languages

\subsubsection{Targeted adaptation}
\begin{table}[t]
\centering
\resizebox{0.48\textwidth}{!}{ 
\begin{tabular}{ r  c c | r  c c}
    \toprule
     \multicolumn{3}{c|}{Asian} & \multicolumn{3}{c}{ME-Indic} \\
     & \textsc{Cluster} & \textsc{Universal} & & \textsc{Cluster} & \textsc{Universal} \\
    \midrule
     arb & 16.5 & 30.5 & deu & 20.5 & 35.5 \\
     deu & 20.0 & 30.5 & ell & 19.1 & 32.5 \\
     ell & 13.1 & 32.0 & fra & 27.5 & 39.0 \\
     fra & 27.5 & 37.0 & ind & 20.0 & 46.5 \\
     heb & 12.1 & 32.0 & jpn & 16.5 & 41.5 \\
     hin &  9.1 & 25.8 & kor & 16.5 & 37.0 \\
     pes & 17.0 & 41.5 & rus & 22.5 & 39.0 \\
     rus & 20.0 & 37.5 & spa & 39.5 & 46.5 \\
     spa & 33.0 & 43.5 & vie & 16.5 & 43.0 \\
     tur & 15.5 & 30.0 & zho & 34.5 & 50.5 \\
    \bottomrule
\end{tabular}%
}
\caption{Targeted language adaptation win rates}
\label{tab:sft-full-results}
\end{table}

Table \ref{tab:cpt-full-results} presents the win rates by language after continued pretraining, divided into \textsc{Primary} and \textsc{Expanded} languages

\subsection{Judge Prompt for Win Rates}
\label{app:winrate-prompt}

\fbox{\begin{minipage}{42em}
\scriptsize
    \textbf{<system\_prompt>} \\
    You are a helpful following assistant whose goal is to select the preferred (least wrong) output for a given instruction. \\
    
    \textbf{<user\_prompt>} \\
    Which of the following answers is the best one for the given instruction? A good answer should follow these rules:\\
    1) It should have correct reasoning,"\\
    2) It should answer the request in the instruction,\\
    3) It should be factually correct and semantically comprehensible,\\
    4) It should be grammatically correct and fluent.\\
    
    Instruction: {instruction}\\
    Answer (A): {completion\_a}\\
    Answer (B): {completion\_b}\\
    
    FIRST provide a concise comparison of the two answers. If one answer is better, explain which you prefer and why. If both answers are identical or equally good or bad, explain why.
    SECOND, on a new line, state exactly one of 'Answer (A)' or 'Answer (B)' or 'TIE' to indicate your choice of preferred response.\\
    Your response should use the format: Comparison: <concise comparison and explanation> Preferred: <'Answer (A)' or 'Answer (B)' or 'TIE'>.     
    \end{minipage}}

\section{Compression Ratio}
\label{app:compression-ratio}
In order to intrinsically evaluate tokenizer quality, we measure compression ratio as compared to the publicly-available multilingual tokenizer used in Command-A \citep{cohere2025commandaenterprisereadylarge}. Compression measures how efficiently data is represented in terms of size (in bytes), and BPE optimizes for this condition \citep{Gage1994ANA}. Compression ratio compares compression values between tokenizers, and since lower compression is desirable, a compression ratio below 1 indicates that a tokenizer has more favorable compression.
Previous work shows that compression correlates well with model performance, especially for generative tasks \citep{goldman-etal-2024-unpacking, galle-2019-investigating}, although lower compression is not a sufficient condition for a better tokenizer \citep{schmidt-etal-2024-tokenization}. However, long sequence lengths are one of the ways in which inequitable treatment of languages begins at the tokenizer \citep{velayuthan2024egalitarianlanguagerepresentationlanguage, ahia-etal-2023-languages}, and is therefore an important measure to consider along with downstream evaluations.

\subsection{Impact of tokenizer language weighting on compression ratio}

\begin{figure}[t]
\centering
   \includegraphics[width=0.46\textwidth]{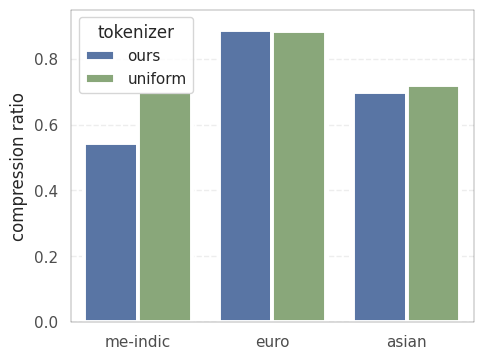}
   \caption{Compression ratios for our tokenizer and the baseline uniform tokenizer. Our tokenizer uses a special weighting that leverages training data distribution and the language grouping (\S~\ref{sec:tokenizer-training}), leading to a better compression (lower is better). }
   \label{fig:compression}
\end{figure}

As a baseline, we evaluate uniform language weighting and compare it with our tokenizer where we use both data distribution and language bucketing strategies in conjunction. Compression ratios are computed against the multilingual tokenizer of open weight Command-A \citep{cohere2025commandaenterprisereadylarge} model on the test split of FineWeb-2 \citep{penedo2024fineweb-2}. 

Figure \ref{fig:compression} shows the comparison. As seen, our tokenizer, which uses a special weighting, leads to better compression than the uniform baseline. Note that both of these tokenizers achieve overall higher compression performance than Command-A since they are trained with larger language coverage.

\subsection{Impact of compression ratio on downstream performance}
\begin{figure}[t]
\centering
   \includegraphics[width=0.42\textwidth]{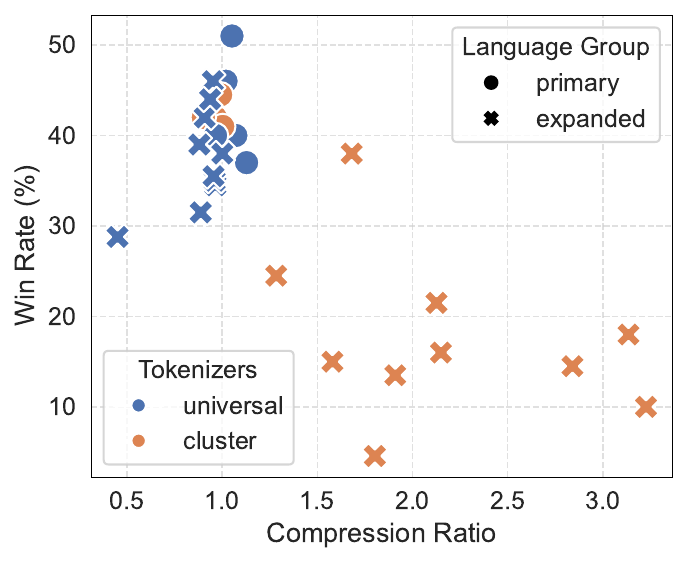}
   \caption{Adaptation results per language in Euro cluster together with tokenizers' compression ratio. While \textsc{Universal} tokenizer enables better compression, especially for the expanded language subset, hence, better downstream performance, \textsc{Cluster} tokenizer fails to represent these languages, leading to lower adaptation results. }
   \label{fig:ratio-vs-winrate}
\end{figure}

Figure \ref{fig:ratio-vs-winrate} explores the relationship between compression ratio and win rates for European cluster models trained with the \textsc{universal} and \textsc{cluster} tokenizer on primary and expanded languages. The expansion languages exhibit large compression ratios with the \textsc{cluster}, all over 1, which indicates that compression in that language is worse than the comparison tokenizer. At the same time, the win rates for those languages are lower than the primary languages, which also have a lower compression ratio. The \textsc{universal} tokenizer, however, shows relatively high win rates and low compression ratios for both primary and expansion languages. This result corroborates the relationship between compression ratio and downstream performance, and provides an additional dimension for language plasticity of the \textsc{universal} tokenizer. 

\section{Languages}
\label{app:language-list}

\scriptsize
\begin{longtable}{l@{\hspace{2pt}}c@{\hspace{2pt}}ccccc@{\hspace{2pt}}c}
\caption{Pretraining languages, including pretraining cluster assignment. Languages with a checkmark in the post-training column but without cluster assignment are used as unseen adaptation languages.}
\label{tab:language_codes} \\
\toprule
ISO Code & Language & Script & Family & Subgrouping & Resources & Cluster & In Post-Training \\
\midrule
\endfirsthead

\toprule
ISO Code & Language & Script & Family & Subgrouping & Resources & Cluster & In Post-Training \\
\midrule
\endhead

\bottomrule
\endfoot

\bottomrule
\endlastfoot
afr & Afrikaans & Latin & Indo-European & Germanic  & Mid & - & \cmark \\
ara & Arabic & Arabic & Afro-Asiatic   & Semitic & High & Me-Indic & \cmark \\
amh & Amharic & Ge'ez & Afro-Asiatic & Semitic & Low & - & \xmark \\
bel & Belarusian & Cyrillic & Indo-European & Balto-Slavic & Mid & - & \cmark \\
ben & Bengali & Bengali & Indo-European  & Indo-Aryan & Mid & Me-Indic & \xmark \\
bul & Bulgarian & Cyrillic & Indo-European  & Balto-Slavic & Mid & Euro & \xmark \\
cat & Catalan & Latin & Indo-European  & Italic & High & Euro & \xmark \\
ces & Czech & Latin & Indo-European & Balto-Slavic & High & Euro & \cmark \\
cym & Welsh & Latin & Indo-European & Celtic & Low & Euro & \xmark \\
dan & Danish & Latin & Indo-European  & Germanic & Mid & Euro & \xmark \\
deu & German & Latin & Indo-European  & Germanic & High & Euro & \cmark \\
ell & Greek & Greek & Indo-European  & Graeco-Phrygian & Mid & Euro & \cmark \\
eng & English & Latin & Indo-European  & Germanic & High & Euro & \cmark \\
est & Estonian & Latin & Uralic & Finnic & Mid & Euro & \xmark \\
eus & Basque & Latin & Basque & - & High & Euro & \xmark \\
fil & Filipino & Latin & Austronesian & Malayo-Polynesian & Mid & Asian & \xmark \\
fin & Finnish & Latin & Uralic & Finnic & Mid & Euro & \xmark \\
fra & French & Latin & Indo-European  & Italic & High & Euro & \cmark \\
gla & Scottish Gaelic & Latin & Indo-European & Celtic & Low & Euro & \xmark \\
gle & Irish & Latin & Indo-European  & Celtic & Low & Euro & \xmark \\
glg & Galician & Latin & Indo-European & Italic & Mid & Euro & \xmark \\
guj & Gujarati & Gujarati & Indo-European  & Indo-Aryan & Low & Me-Indic & \xmark \\
heb & Hebrew & Hebrew & Afro-Asiatic & Semitic & Mid & Me-Indic & \cmark \\
hin & Hindi & Devanagari & Indo-European  & Indo-Aryan & High & Me-Indic & \cmark \\
hrv & Croatian & Latin & Indo-European & Balto-Slavic & High & Euro & \xmark \\
hun & Hungarian & Latin & Uralic & - & High & Euro & \xmark \\
hye & Armenian & Armenian & Indo-European &  Armenic & Low & - & \cmark \\
ibo & Igbo & Latin & Atlantic-Congo & Benue-Congo & Low & - & \xmark \\
ind & Indonesian & Latin & Austronesian & Malayo-Polynesian & Mid & Asian & \cmark \\
ita & Italian & Latin & Indo-European  & Italic & High & Euro & \cmark \\
jav & Javanese & Latin & Austronesian & Malayo-Polynesian & Low & Asian & \xmark \\
jpn & Japanese & Japanese & Japonic & Japanesic & High & Asian & \cmark \\
kaz & Kazakh & Cyrillic & Turkic & Common Turkic & Mid & - & \cmark \\ 
khm & Khmer & Khmer & Austroasiatic & Khmeric & Low & Asian & \xmark \\
kor & Korean & Hangul & Koreanic & Korean & Mid & Asian & \cmark \\
lao & Lao & Lao & Tai-Kadai & Kam-Tai & Low & Asian & \xmark \\
lav & Latvian & Latin & Indo-European & Balto-Slavic & Mid & Euro & \xmark \\
lit & Lithuanian & Latin & Indo-European  & Balto-Slavic & Mid & Euro & \xmark \\
mlt & Maltese & Latin & Afro-Asiatic & Semitic & Low & Me-Indic & \xmark \\
msa & Malay & Latin & Austronesian & Malayo-Polynesian & Mid & Asian & \xmark \\
mya & Burmese & Myanmar & Sino-Tibetan & Burmo-Qiangic & Low & Asian & \xmark \\
nep & Nepali & Devanagari & Indo-European  & Indo-Aryan & Low & - & \cmark\\
nld & Dutch & Latin & Indo-European  & Germanic & High & Euro & \cmark \\
nor & Norwegian & Latin & Indo-European  & Germanic & Low & Euro & \xmark \\
pan & Punjabi & Gurmukhi & Indo-European  & Indo-Aryan & Low & Me-Indic & \xmark \\
pes & Persian & Arabic & Indo-European  & Iranian & High & Me-Indic & \cmark \\
pol & Polish & Latin & Indo-European  & Balto-Slavic & High & Euro & \cmark \\
por & Portuguese & Latin & Indo-European  & Italic & High & Euro & \cmark \\
ron & Romanian & Latin & Indo-European & Italic & Mid & Euro & \cmark \\
rus & Russian & Cyrillic & Indo-European & Balto-Slavic & High & Euro & \cmark \\
sin & Sinhala & Sinhala & Indo-European  & Indo-Aryan & Low & - & \cmark \\
slk & Slovak & Latin & Indo-European  & Balto-Slavic & Mid & Euro & \xmark \\
slv & Slovenian & Latin & Indo-European  & Balto-Slavic & Mid & Euro & \xmark \\
spa & Spanish & Latin & Indo-European  & Italic & High & Euro & \cmark \\
srp & Serbian & Cyrillic & Indo-European  & Balto-Slavic & High & Euro & \xmark \\
swa & Swahili & Latin & Atlantic-Congo & Benue-Congo & Low & - & \xmark \\
swe & Swedish & Latin & Indo-European  & Germanic & High & Euro & \xmark \\
tam & Tamil & Tamil & Dravidian & South Dravidian & Mid & Me-Indic & \xmark \\
tel & Telugu & Telugu & Dravidian & South Dravidian & Low & Me-Indic & \xmark \\
tha & Thai & Thai & Tai-Kadai & Kam-Tai & Mid & Asian & \xmark \\
tur & Turkish & Latin & Turkic & Common Turkic & High & Me-Indic & \cmark \\
ukr & Ukrainian & Cyrillic & Indo-European  & Balto-Slavic & Mid & Euro & \cmark \\
urd & Urdu & Arabic & Indo-European  & Indo-Aryan & Mid & Me-Indic & \xmark \\
vie & Vietnamese & Latin & Austroasiatic & Vietic & High & Asian & \cmark \\
xho & Xhosa & Latin & Atlantic-Congo & Benue-Congo & Low & - & \xmark \\
yor & Yorùbá & Latin & Atlantic-Congo & Benue-Congo & Low & - & \xmark \\
yue & Cantonese & Han & Sino-Tibetan & Sinitic & Low & - & \cmark \\
zho & Mandarin Chinese & Han & Sino-Tibetan & Sinitic & High & Asian & \cmark \\
zul & Zulu & Latin & Atlantic-Congo & Benue-Congo & Low & - & \xmark \\
\end{longtable}

\end{document}